\documentclass[runningheads]{llncs}

\usepackage{eccv}

\usepackage{eccvabbrv}

\usepackage{graphicx}
\usepackage{booktabs}

\usepackage[accsupp]{axessibility}  

\usepackage{hyperref}

\usepackage{orcidlink}

\begin{document}

\title{Uncertainty Estimation and Out-of-Distribution Detection for LiDAR Scene Semantic Segmentation} 

\titlerunning{Uncertainty Estimation and Out-of-Distribution Detection}

\author{Hanieh Shojaei Miandashti\inst{1}\orcidlink{0000-0003-0968-0420} \and
Qianqian Zou\inst{1}\orcidlink{0000-0001-7673-0630} \and
Max Mehltretter\inst{2}\orcidlink{0000-0002-3708-9868}
}

\authorrunning{H. Shojaei et al.}

\institute{Institute of Cartography and Geoinformatics, Leibniz University Hannover, Germany 
\email{\{hanieh.shojaei,qianqian.zou\}@ikg.uni-hannover.de} \and
Institute of Photogrammetry and GeoInformation, Leibniz University Hannover, Germany
\email{mehltretter@ipi.uni-hannover.de}}

\maketitle

\begin{abstract}
 Safe navigation in new environments requires autonomous vehicles and robots to accurately interpret their surroundings, relying on LiDAR scene segmentation, out-of-distribution (OOD) obstacle detection, and uncertainty computation. We propose a method to distinguish in-distribution (ID) from OOD samples and quantify both epistemic and aleatoric uncertainties using the feature space of a single deterministic model. After training a semantic segmentation network, a Gaussian Mixture Model (GMM) is fitted to its feature space. OOD samples are detected by checking if their squared Mahalanobis distances to each Gaussian component conform to a chi-squared distribution, eliminating the need for an additional OOD training set. Given that the estimated mean and covariance matrix of a multivariate Gaussian distribution follow Gaussian and Inverse-Wishart distributions, multiple GMMs are generated by sampling from these distributions to assess epistemic uncertainty through classification variability. Aleatoric uncertainty is derived from the entropy of responsibility values within Gaussian components. Comparing our method with deep ensembles and logit-sampling for uncertainty computation demonstrates its superior performance in real-world applications for quantifying epistemic and aleatoric uncertainty, as well as detecting OOD samples. While deep ensembles miss some highly uncertain samples, our method successfully detects them and assigns high epistemic uncertainty.
  
\end{abstract}

\section{Introduction}
\label{sec:intro}

Despite the capability of advanced deep learning models to accurately assign semantic labels to LiDAR point clouds, there is a notable lack of methods for uncertainty quantification. However, the estimation of uncertainty is essential for assessing the reliability of any prediction, particularly for safety-critical systems such as autonomous vehicles that rely on real data, including LiDAR point clouds. These systems need not only to perceive their surroundings but also to quantify uncertainty to avoid over-reliance on potentially erroneous predictions. Two primary types of uncertainty are generally distinguished: epistemic and aleatoric. Epistemic uncertainty, which arises from the model itself, reflects the reliability of a model's predictions, whereas aleatoric uncertainty stems from characteristics inherent in the data \cite{kendall2017}.

Current deep learning-based methods for quantifying epistemic uncertainty typically require extensive retraining of a model for sampling from the posterior distribution \cite{gal2016, lakshminarayanan2017}, rendering them slow and unsuitable for real-time applications, or they fail to effectively distinguish between epistemic and aleatoric uncertainties. Furthermore, aleatoric uncertainty is commonly measured using the entropy of softmax outputs from a deep network. However, deep discriminative classifiers often exhibit overconfidence in their uncalibrated softmax predictions, which can lead to misleading estimates of aleatoric uncertainty \cite{guo2017, gawlikowski2023}.

To address the mentioned limitations in uncertainty quantification, we propose training a discriminative deep network for semantic segmentation, which focuses on decision boundaries between classes rather than modeling the full distribution of features for each class. However, to quantify both epistemic and aleatoric uncertainty, we subsequently fit a GMM with a component for each class to the high-dimensional feature space of this deterministic model, capturing the overall distribution of the training data features.

We introduce a consistent metric based on the probabilistic uncertainty associated with the estimated GMM distribution in feature space. Considering that the estimated mean and covariance matrix of a multivariate normal distribution adhere to Gaussian and Inverse-Wishart distributions, respectively \cite{bishop2006}, we advance our methodology by modeling distributions over these Gaussian parameters. This enables us to sample multiple GMMs from these distributions, facilitating a spectrum of density evaluations for a test sample rather than a single deterministic density derived from one GMM configuration. We define our measure of epistemic uncertainty based on the entropy of the classification outcomes across these sampled GMMs. Concurrently, aleatoric uncertainty is robustly quantified by computing the entropy of the responsibility values obtained from the sampled GMMs.

Regarding the OOD detection, unlike methods relying on additional OOD dataset to determine a threshold to distinguish between ID and known OOD samples \cite{malinin2018}, we employ a Chi-square test on the Mahalanobis distance between a sample and each component of the GMM to identify OOD samples.

To summarize, we propose a novel uncertainty estimation and OOD detection method for semantic segmentation of LiDAR point clouds that is adaptable to practical, real-world applications. The primary contributions are as follows:

\begin{enumerate}

    \item \textbf{Generative-Discriminative Approach:} We employ a hybrid strategy by first training a discriminative classifier, which requires fewer parameters and focuses on learning decision boundaries for the classification task efficiently. Then, we integrate a generative model (GMM) in the feature space to capture the full distribution of class features, potentially leading to better calibrated and more robust uncertainty estimations.
   
    \item \textbf{Epistemic Uncertainty Estimation:} We propose a consistent epistemic uncertainty metric that leverages Gaussian and Inverse-Wishart distributions on the empirical mean and covariance matrix of GMM components to generate multiple generative classifiers. We compute the entropy of the frequency distribution of classification outcomes, providing a robust measure of epistemic uncertainty that captures the variability across these sampled models.
    
    \item \textbf{Aleatoric Uncertainty Estimation:} We introduce the entropy of the calibrated responsibility values from the GMM, considering the entire data distribution in a high-dimensional feature space, as a measure of aleatoric uncertainty.

    \item \textbf{Out-of-distribution Detection:} We employ the Mahalanobis distance coupled with a Chi-square test to robustly identify OOD samples. Consequently, our method does not require OOD samples during training to learn their detection.

\end{enumerate}

\section{Related Works}

Two main approaches for quantifying epistemic uncertainty are sampling-based methods and single-forward pass methods. Monte Carlo Dropout \cite{gal2016} and deep ensembles \cite{lakshminarayanan2017} are notable sampling-based methods, with deep ensembles generally offering superior performance \cite{ovadia2019}. In deep ensembles, multiple network realizations are trained with randomness, such as different initializations, and epistemic uncertainty is estimated by the variance of their predictions. However, this approach requires training multiple models. In contrast, single-forward pass models have gained attention recently for computing epistemic uncertainty without the need for multiple iterations or models \cite{van2020, liu2020, lee2018}.


To compute uncertainty and to identify OOD samples in a single deterministic model, Lee et al. replaced the softmax layer with Gaussian Discriminative Analysis (GDA) \cite{lee2018}. This method calculates a confidence score based on the Mahalanobis distance to the nearest class distribution. However, it suffers from feature collapse, where OOD inputs are mapped to ID areas, making them indistinguishable \cite{van2021}. To address this limitation, DUQ \cite{van2020} and SNGP \cite{liu2020} use distance-aware output layers with radial basis functions (RBFs) and Gaussian processes (GPs), incorporating inductive biases through a Jacobian penalty or spectral normalization, respectively. Although effective, these methods significantly alter the training process, add hyper-parameters, and cannot separate aleatoric from epistemic uncertainty, which is crucial for certain applications, such as OOD detection and active learning \cite{nguyen2022}.

Deep Deterministic Uncertainty (DDU) \cite{mukhoti2023} computes epistemic and aleatoric uncertainties separately without modifying the standard deep learning model. Using spectral normalization, DDU prevents feature collapse by ensuring smoothness in the feature extractor. It fits a GMM with one component per class to the feature space and distinguishes between ID and OOD samples by analyzing GMM density probabilities. 

DDU computes epistemic uncertainty based on GMM density, with lower density indicating higher uncertainty. However, this approach can be inconsistent across different problems, datasets, and feature spaces due to significant variability in density values. Additionally, DDU lacks a robust method for quantifying aleatoric uncertainty, relying on discriminative classifier outputs that often lead to overconfident predictions \cite{guo2017, gawlikowski2023}. 


Aleatoric uncertainty is computed through various methods focusing on the distribution over a model’s outputs \cite{kendall2017}. These include noise propagation \cite{loquercio2020}, logit-sampling \cite{kendall2017}, and Shannon entropy of softmax outputs \cite{hullermeier2021}. Logit-sampling assumes two outputs per class in the last layer, representing the mean and variance over the output distribution. The results of this approach have been shown to be better-calibrated than softmax probabilities for computing aleatoric uncertainty \cite{kendall2017, dreissig2023}. We compare aleatoric uncertainties from our method with those from logit-sampling.

\section{Methodology}
\begin{figure}
  \centering
  \includegraphics[width=1\textwidth]{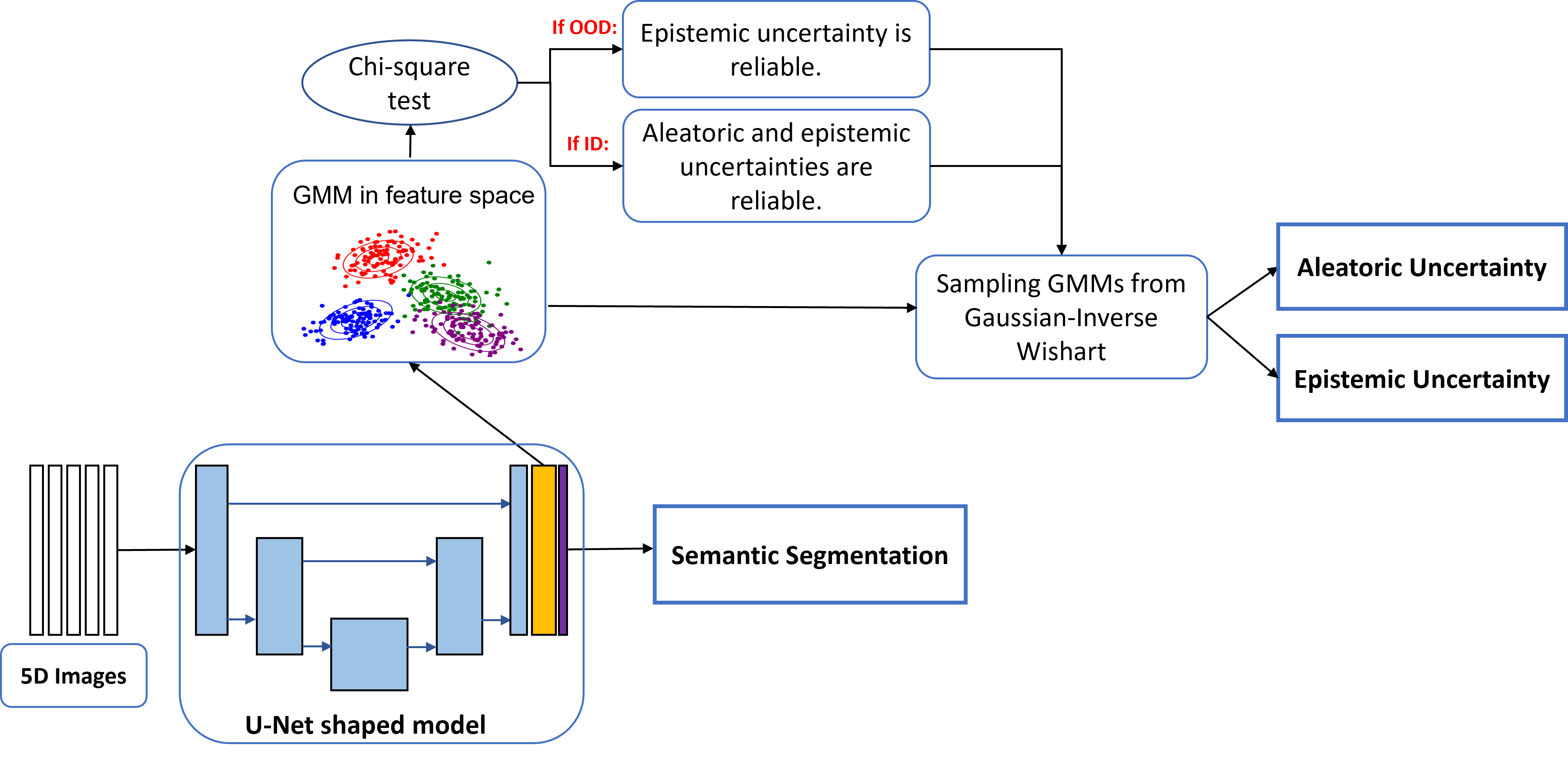}  
  \caption{Overview of our proposed method for OOD detection and epistemic and aleatoric uncertainty estimation. In this work, SalsaNext \cite{cortinhal2020} is used as \emph{U-Net shaped model}, but other models designed for semantic segmentation could be used as well. The outputs of the layer colored in yellow are selected as the feature space to which a GMM is fitted and the layer colored in purple represents the classification layer producing the softmax outputs.}
  \label{fig:Flowchart}
\end{figure}

In this section, we detail our methodology, illustrated in \cref{fig:Flowchart}, which begins by projecting a 3D point cloud into a 2D spherical range-view image\cite{milioto2019}. This results in a 2D image encompassing five channels, containing the point coordinates (x, y, z), intensity, and range values from each point in the LiDAR point cloud. This 5D range-view image is subsequently fed into a U-Net shaped model named SalsaNext \cite{cortinhal2020}, which is designed to predict the class label for each pixel of the input range-view image.

After training, a distribution over GMMs is modelled using Gaussian and Inverse-Wishart distributions in the feature space of the last convolutional layer. 
The Gaussian parameters are estimated probabilistically, instead of simple point estimates. Thus, we can sample an ensemble of GMMs from the distribution of parameters. 
Given test data, the probability density of each class is computed for classification across all sampled GMMs. The epistemic uncertainty is derived from the entropy of the averaged classification outcomes across all GMM samples, while aleatoric uncertainty is obtained from the entropy of responsibilities of different Gaussian components. 

We differentiate ID and OOD samples by checking if the squared Mahalanobis distance from a test sample to any Gaussian component follows a chi-squared distribution. For ID samples, both epistemic and aleatoric uncertainties are quantified, while for OOD samples, only epistemic uncertainty is valid. 

\subsection{Fitting GMM to the Feature Space}
Following the training of the model, described later in \cref{Data Preparation and Semantic Segmentation}, a GMM is fitted to the feature space, with the number of components (\( C \)) corresponding to the number of classes, each operating in a \( d \)-dimensional feature space. We postulate, based on the findings from \cite{mukhoti2023}, that samples from each class can be reasonably well approximated by a multivariate Gaussian distribution, the dimensions of which are equal to those of the feature space. To fit this GMM, empirical means (\( \{ \hat{\mu}_c \}_{c=1}^C \)) and covariance matrices (\( \{ \hat{\Sigma}_c \}_{c=1}^C \)) are calculated for each class across the entire training dataset. Additionally, the frequency of each class is utilized as the prior weight for the corresponding GMM component (\( \{ \pi_c \}_{c=1}^C \)).

\subsection{Epistemic Uncertainty}
Considering that Gaussian and Inverse-Wishart distributions serve as distributions over the estimated mean and covariance matrix of a multivariate normal distribution, they are well-suited for modeling uncertainty in the GMM parameters.
In our proposed method, rather than obtaining point estimates for the mean and covariance of each component in the GMM, we adopt the probabilistic estimation of these GMM parameters by utilizing Gaussian and Inverse-Wishart distributions. Consequently, this allows us to sample an ensemble of GMMs from the distributions of these parameters. The estimation process for this distribution of GMMs, followed by the quantification of epistemic uncertainty, proceeds as outlined below:

\begin{itemize}
    \item \textbf{Mean:}
    We assume that each estimated mean \( \hat{\mu}_c \) follows a Gaussian distribution. Specifically, we use \( \mathcal{N}(\hat{\mu}_c \mid \hat{\mu}_c, \Sigma'_c) \), where \( \Sigma'_c \) is the covariance matrix that reflects the uncertainty in the mean estimate.

    \item \textbf{Covariance:}
    We assume that each estimated covariance matrix \( \hat{\Sigma}_c \) follows an Inverse-Wishart distribution. Specifically, we use \( \text{Inverse-Wishart}(\hat{\Sigma}_c \mid \Psi_c, \nu) \), where \( \Psi_c \) is the scale matrix derived from the empirical data and \( \nu \) is the degrees of freedom.  

     \item \textbf{Sampling:}
     We draw \( T \) samples from these distributions:

    \begin{itemize}
        \item Sample \( \{ \tilde{\mu}_{c,t} \}_{t=1}^T \) from \( \mathcal{N}(\hat{\mu}_c \mid \hat{\mu}_c, \Sigma'_c) \).
        \item Sample \( \{ \tilde{\Sigma}_{c,t} \}_{t=1}^T \) from \( \text{Inverse-Wishart}(\hat{\Sigma}_c \mid \Psi_c, \nu) \).
    \end{itemize}
    
    This generates \( T \) GMMs, each defined by the parameters \( \{ \tilde{\mu}_{c,t}, \tilde{\Sigma}_{c,t} \}_{t=1}^T \).

    \item \textbf{Epistemic Uncertainty Quantification:}
    For each sampled GMM, we classify a data point by comparing the probability density values derived from \cref{eq:Probability_Density} and then compute the frequency with which each data point is assigned to each class across all \( T \) GMMs. Specifically, for a data point \( x \), let \( f_{c} \) represent the frequency of assigning \( x \) to class \( c \) across the \( T \) GMMs.

    \begin{equation}
        p(x) = \sum_{c=1}^C \pi_c \, \mathcal{N}(x \mid \tilde{\mu}_c, \tilde{\Sigma}_c),
        \label{eq:Probability_Density}
    \end{equation}

    For each data point \( x \), we compute the entropy of the frequency distribution \( \{ f_{c} \}_{c=1}^C \), where \( C \) is the number of classes. The entropy of the frequency distribution over the classes (\( H(f) \)) for each data point is given by:
    
    \begin{equation}
    H(f) = -\sum_{c=1}^C f_{c} \log f_{c}
    \label{eq:EU}
    \end{equation}

    The entropy values \( H(f) \) provide a measure of epistemic uncertainty. Higher entropy indicates greater uncertainty in classification, reflecting the variability in class assignments across sampled GMMs. This variability stems from uncertainty in the estimated parameters of the generative model (GMM) used to represent the data distribution in the feature space.

\end{itemize}

\subsection{Aleatoric Uncertainty}
To approximate aleatoric uncertainty, we propose calculating the mean of responsibility values for each sample derived from \( T \) sampled GMMs. Responsibility values for a sample \( x \) from a sampled GMM is defined by \cref{eq:Responsibility_value}. In this context, \( \tilde{\gamma}_t(x_{c}) \) denotes the probability that a sample \( x \) originates from the \( c \)-th Gaussian component \cite{bishop2006}.  

\begin{equation}
    \tilde{\gamma}_t(x_{c}) = \frac{\pi_c \, \mathcal{N}(x \mid \tilde{\mu}_{c,t} \, \tilde{\Sigma}_{c,t})}{\sum_{c=1}^C \pi_c \, \mathcal{N}(x \mid \tilde{\mu}_{c,t}, \tilde{\Sigma}_{c,t})}
    \label{eq:Responsibility_value}
\end{equation}

We employ the entropy of the responsibility values to approximate the aleatoric uncertainty, which is defined in \cref{eq:Entropy_RV}. In this equation, \( \gamma(x_{c}) \) is the mean of \( \tilde{\gamma_t}(x_{c}) \) across $T$ sampled GMMs. This measure is low when a single multivariate Gaussian component, corresponding to one class, clearly includes the sample, and it is high if multiple Gaussians overlap and share the responsibility, indicating that this sample cannot unambiguously be assigned to a specific class.

\begin{equation}
    H(\gamma(x)) = -\sum_{c=1}^C \gamma(x_{c}) \log \gamma(x_{c})
    \label{eq:Entropy_RV}
\end{equation}

\subsection{OOD Detection}
For multivariate Gaussian data, the distribution of the squared Mahalanobis distance is known to be chi-squared (\( \chi^2_d \)) with $d$ (the dimension of the data) degrees of freedom \cite{gnanadesikan1972}. To identify OOD samples, we test whether a sample (\( x \)) falls within the 0.975 quantiles of the Chi-square distribution (\( \chi^2_d,0.025 \)) \cite{maronna2002}. The Mahalanobis distance (\( D(x)\)) of each sample to the Gaussian components is computed to determine whether a sample is considered as ID or OOD: 

\textbf{ID:} If \( D(x)^2 \leq \chi^2_{\text{0.025}} \), the sample is considered to be ID.

\textbf{OOD:} If \( D(x)^2 > \chi^2_{\text{0.025}} \), the sample is considered to be OOD.

\noindent OOD samples, are considered unreliable for determining aleatoric uncertainty, as the model has not encountered any related samples during training. 
Therefore, just epistemic uncertainty is assumed to be reliable for these samples.

\section{Experimental Setup}
\label{sec:blind}
In this section, we detail the technical aspects of the deep learning model used for semantic segmentation and the application of a GMM to the feature space of our trained model. 
 
\subsection{Data Preparation and Semantic Segmentation}
\label{Data Preparation and Semantic Segmentation}
We utilized the SemanticKITTI dataset \cite{behley2019} as a benchmark for LiDAR scene semantic segmentation, which provides a semantic class label for each point. Each 3D scan is transformed into 2D spherical range-view images \cite{milioto2019} with a size of [64 × 2048] pixels. The 3D point coordinates (x, y, z), intensity values, and range values are stored across separate range-view image channels. This transformation produces an image with dimensions of [64 × 2048 × 5] pixels, which serves as the input for the SalsaNext network \cite{cortinhal2020}, which performs semantic segmentation on this data, categorizing it into 20 object classes commonly found in urban scenes. Notably, we excluded the outliers class defined in the dataset from the training process, treating it as OOD since the model does not encounter it during training. We trained SalsaNext using the Adam optimizer and an equal-weighted sum of cross-entropy and Lovasz softmax \cite{berman2018} as the loss function over 100 epochs. The model was trained on sequences 0 to 10, allocating 85\% of the data for training and 15\% for validation, excluding sequence 8, which was reserved for testing. Early stopping was used to prevent overfitting and to optimize model performance. 

\subsection{Fitting GMM to the Feature Space}
After training, the 32-dimensional layer before the classifier is selected as the feature space. Assuming the features of training samples across 20 classes are well approximated by a 32-dimensional multivariate Gaussian distribution, we use the Gaussian Inverse-Wishart distribution over the estimated means (32) and covariance matrices (32x32) of each class to derive 20 \( T \) sampled GMMs for further analysis.

\subsection{Evaluation Metrics}
To assess the effectiveness of our uncertainty estimation method, we employed confidence histograms and the Expected Calibration Error (ECE) \cite{guo2017} to evaluate the estimated uncertainties. Additionally, to gauge the performance of the OOD detection method, we calculated the confusion matrix, which considers both ID and OOD samples as identified by our approach against the ground truth.

\section{Experimental Results}

\subsection{OOD Detection} 
\begin{figure}
    \centering
    \begin{subfigure}{0.95\textwidth}
        \centering
        \includegraphics[width=\textwidth]{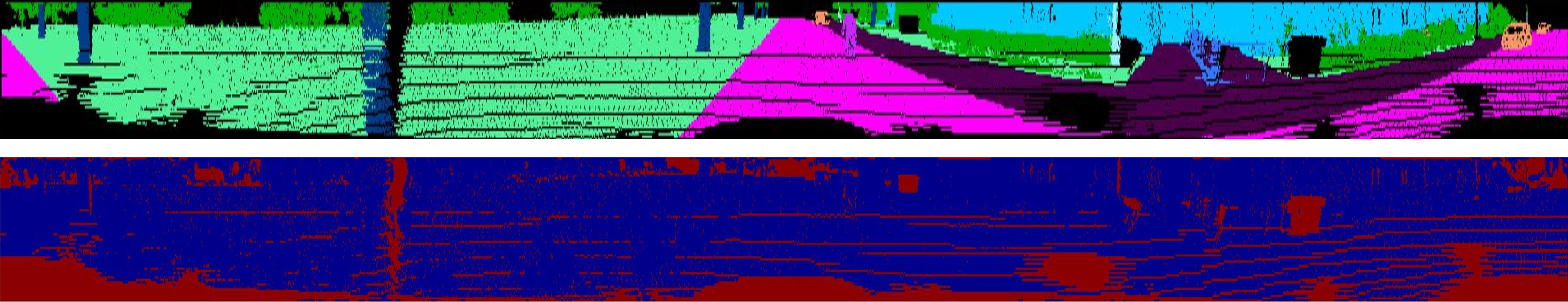}
        \caption{}
        \label{fig:206_OOD}
    \end{subfigure}
    \vfill
    \begin{subfigure}{0.95\textwidth}
        \centering
        \includegraphics[width=\textwidth]{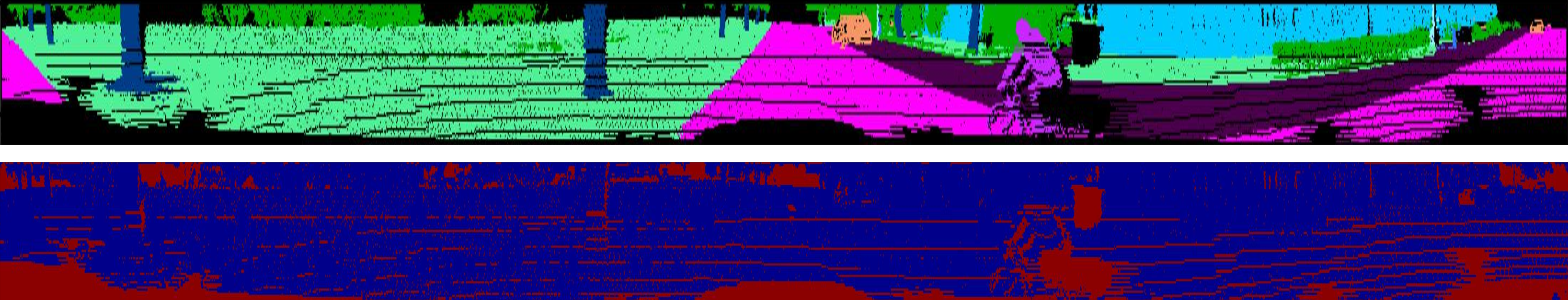}
        \caption{}
        \label{fig:218_OOD}
    \end{subfigure}
    \caption{OOD detection using our proposed method across two different scans. From top to bottom for each 
    \emph{(a)} and \emph{(b)}: \emph{ground truth map}, with OOD samples marked in black and \emph{OOD prediction map}, illustrating OOD samples in red and ID samples in blue color.}
    \label{fig:OOD}
\end{figure}

OOD samples are detected by identifying those with squared Mahalanobis distances that do not conform to a chi-square distribution at a 0.975 significance level. We evaluate our proposed method both quantitatively—using precision, recall, and F1 score metrics—and qualitatively, as shown in \cref{fig:OOD}.

In \cref{fig:OOD}, OOD samples across two scans are depicted. In the ground truth map (first map), OOD samples are distinctly marked in black, differentiating them from other classes shown in color. Our proposed detection method identifies and highlights these OOD samples in red on the second map. To quantitatively evaluate our method, we computed the mean precision, recall, and F-score for OOD detection across the entire test data. The results are as follows: precision = 0.9667, recall = 0.9760, and F-score = 0.9713. The corresponding confusion matrix, normalized per column to show the percentage of true OOD samples confused with true ID samples, is shown in \cref{tab:Confusion_Matrix}. 

\begin{table}[tb]
  \caption{Confusion matrix for detecting OOD samples.}
  \label{tab:Confusion_Matrix}
  \centering
  \begin{tabular}{@{}lll@{}}
    \toprule
    & True OOD & True ID\\
    \midrule
    Predicted OOD  &  97.60\% & 3.36\%\\
    Predicted ID &  2.40\% & 96.64\%\\
  \bottomrule
  \end{tabular}
\end{table}
This table indicates that 97.60\% of true OOD samples were correctly identified, while 2.40\% have been misclassified as ID samples after being projected into the ID area in the feature space. This suggests that spectral normalization has effectively prevented feature collapse. It is anticipated that misclassified OOD samples exhibit higher epistemic uncertainty as they fall in the tails of the Gaussian distribution far from the mean.

\subsection{Uncertainty Quantification}
\label{sec:uncertainty_quantification}

Epistemic and aleatoric uncertainties are quantified by the entropy of frequency of class assignments among multiple sampled GMMs for ID and OOD samples and the entropy of the mean responsibility values within Gaussian components for ID samples, respectively. 

\begin{figure}
  \centering
  \includegraphics[width=0.90\textwidth]{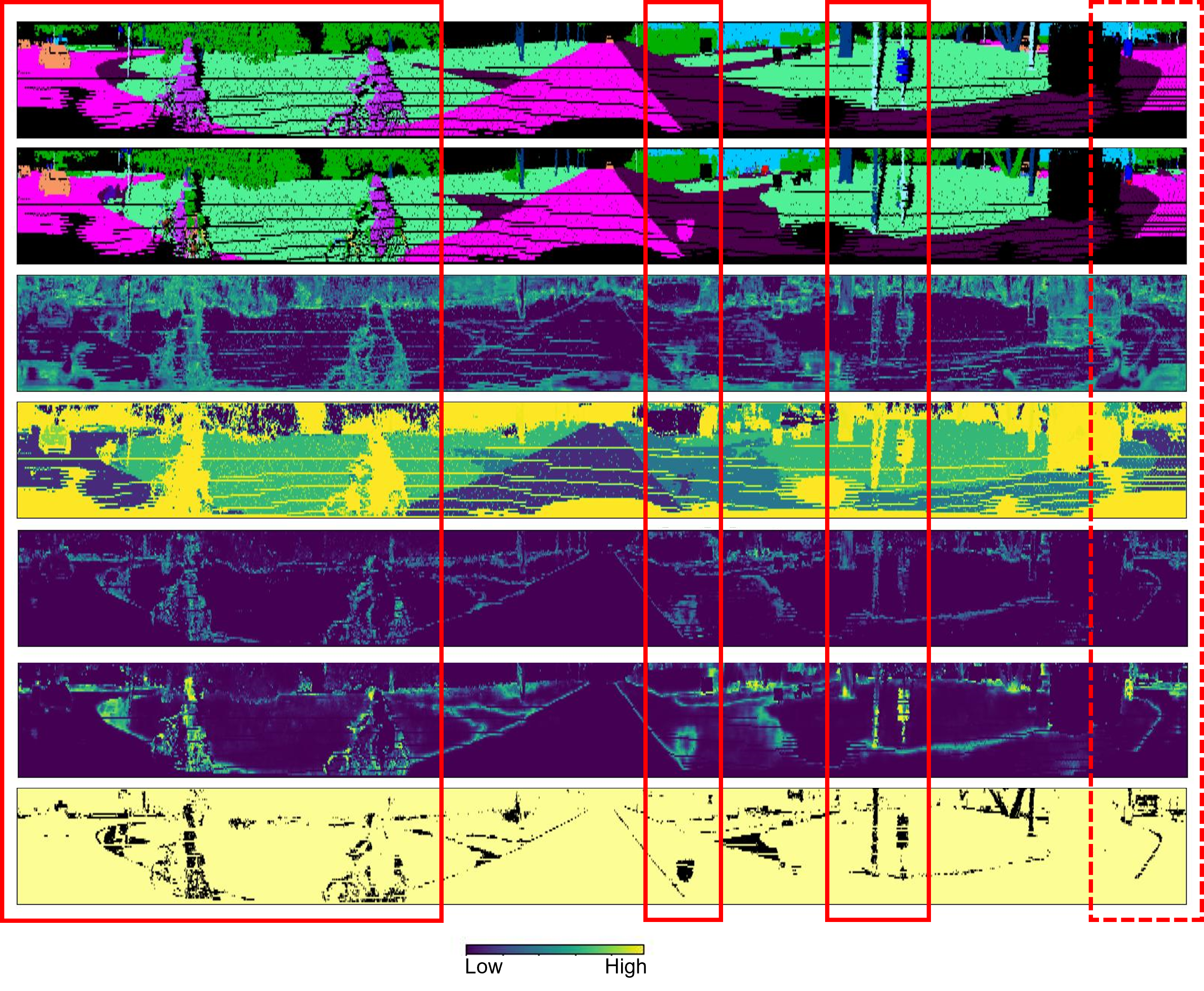} 
  \caption{Epistemic and aleatoric uncertainty quantification by our proposed method and the DDU approach. From top to bottom: \emph{Ground truth map}, \emph{predicted semantic labels}, \emph{our proposed epistemic uncertainty map} derived from the entropy of classification outputs across sampled GMMs, \emph{Epistemic map from DDU} derived from the probability density of a single GMM with empirical means and covariance matrices, \emph{Our proposed aleatoric uncertainty map} by the entropy of responsibility values, \emph{aleatoric map from DDU} by the entropy of softmax outputs, and the \emph{Error map} showing erroneous predictions in black. The highest epistemic uncertainty values are observed in OOD samples and misclassified objects, as highlighted in the red boxes around parts of the bicyclist, the ground beneath the car, the misclassified sidewalk, and the often-confused pole and trunk. The highest aleatoric uncertainty values are observed along class borders and in distant objects.}
  \label{fig:OOD-Uncert}
\end{figure}

\cref{fig:OOD-Uncert} represents epistemic and aleatoric uncertainties computed by our proposed method and by the DDU method.

In \cref{fig:OOD-Uncert}, terrains and streets colored dark green and dark pink in the ground truth (first map) are well classified in the predicted map (second map). Given the large number of samples in these regions, the model is expected to be highly certain about these classes in terms of epistemic uncertainty. This is confirmed in our proposed epistemic uncertainty map (third map), where both classes exhibit low uncertainty. In contrast, the DDU approach (fourth map) shows low uncertainty on the street but higher uncertainty on the terrain areas on the right and left sides of the scan. In all uncertainty maps, more yellowish colors indicate greater uncertainty.

Our proposed epistemic uncertainty approach assigns high uncertainty to specific parts of an object that are not confidently classified, rather than the entire object. In the left red box of \cref{fig:OOD-Uncert}, DDU predicts high uncertainty for all bicycles (yellow), bicyclists (light pink), and cars (orange). In contrast, our method assigns high uncertainty only to the uncertain parts of the bicyclists and bicycles and the ground beneath the car which might have been confused with the car during labeling. Correctly classified portions show lower epistemic uncertainty by our method. Other high-uncertainty areas include a misclassified sidewalk (dark purple, middle red box) and poles/trunk (light/dark blue, right red box) often confused due to their similarities. It is noteworthy that OOD samples (black) exhibit the highest epistemic uncertainty in our method.
        
Aleatoric uncertainty is highest along class boundaries, on distant objects, or in occluded areas where features may not align with the Gaussian component of their class, leading to high entropy in responsibility values. The aleatoric uncertainty maps from our proposed method (fifth map) and the DDU approach (sixth map) closely align with the error map, showing high uncertainties at boundaries and on misclassified objects. However, the DDU approach generally shows higher uncertainty along borders, often extending over a broader margin, making it less confident on objects near boundaries that the model can otherwise distinguish well. Furthermore, while DDU exhibits high uncertainty for certain distant objects that are clearly classifiable, such as the traffic sign outlined in the dashed red box, our proposed method for quantifying aleatoric uncertainty does not exhibit this limitation.

\subsection{Comparison with Existing Methods}

\subsubsection{Comparison of Epistemic Uncertainty and OOD Detection Computed by Our Method, DDU and Deep Ensembles}

\begin{figure}
    \centering
  \includegraphics[width=0.90\textwidth]{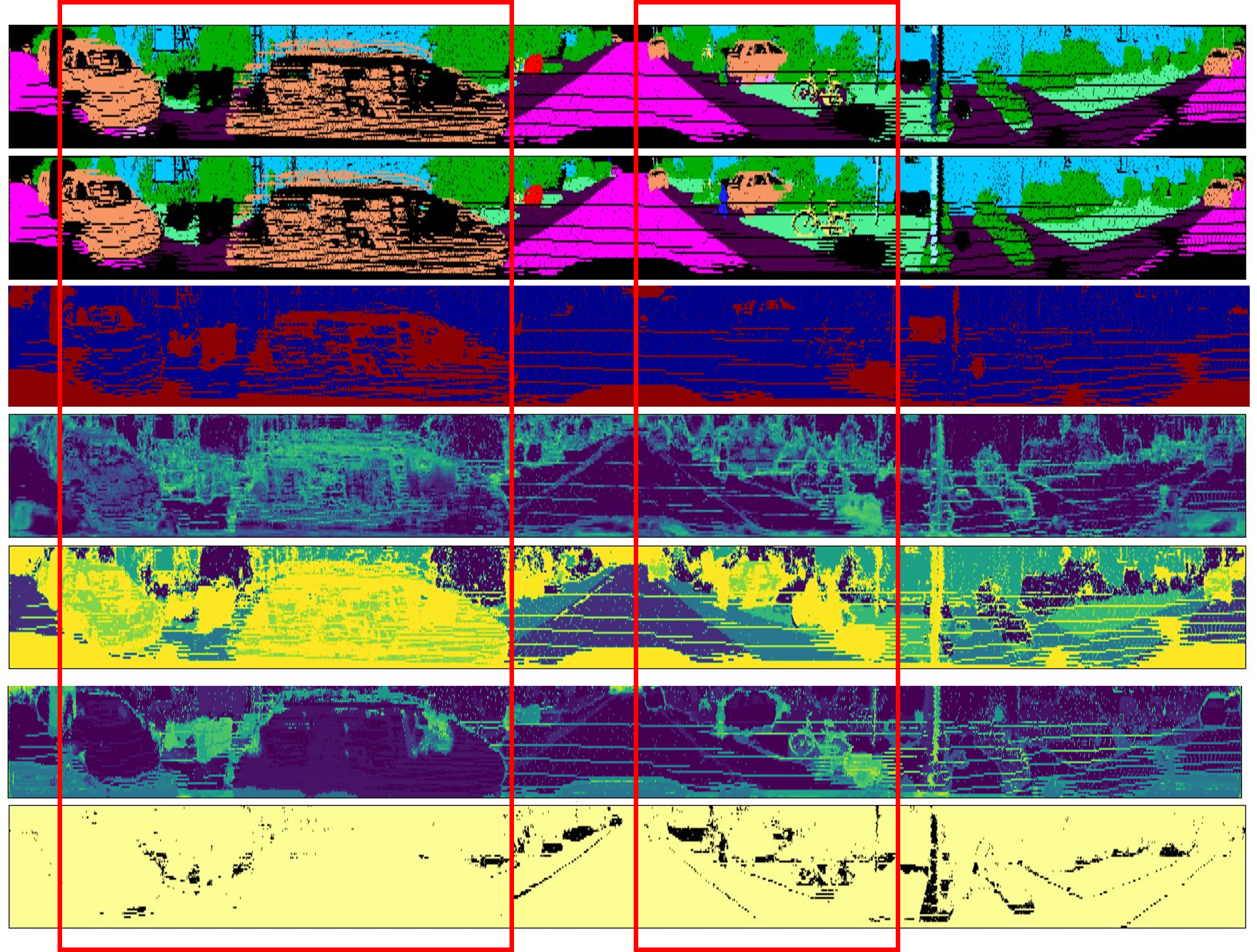}  
  \caption{Comparison of epistemic uncertainties computed by our proposed method, DDU approach and deep ensembles. From top to bottom: \emph{ground truth map}, \emph{predicted semantic labels}, \emph{OOD prediction map by our method}, \emph{eepistemic map by our method}, \emph{epistemic map by DDU}, \emph{epistemic map by deep ensembles}, and the \emph{Error map}. OOD samples are shown in red in the \emph{OOD prediction map} and in black in the \emph{Ground truth map}. Red boxes highlight the uncertain samples that our proposed method successfully detected, which were missed by the deep ensembles and DDU approaches.}

  \label{fig:EU_Comparison}
\end{figure}

Regarding OOD detection, our proposed chi-square test approach clearly distinguishes OOD samples from ID samples, as shown in the third map of \cref{fig:EU_Comparison}. The epistemic uncertainty map from the deep ensembles algorithm (sixth map) highlights high uncertainty for most OOD samples. However, certain OOD samples, like those within cars (red box, left side), do not show significant uncertainty with deep ensembles but are highly uncertain in our proposed map (fourth map) and the DDU epistemic map (fifth map). Additionally, our method effectively detects OOD samples without relying on epistemic uncertainty values, unlike the deep ensembles approach, which requires deciding on a threshold on the epistemics to distinguish OODs from the other uncertain samples.

In addition to better highlighting OOD samples with high epistemic uncertainty, our proposed method provides more detailed uncertainty on specific parts of an object, unlike DDU, which considers the entire object as highly uncertain, as shown for the bicycle and cars in the red box on the right. In this case, both deep ensembles and our approach outperformed DDU, demonstrating the superior capability of our method for epistemic uncertainty compared to deep ensembles and the DDU method.

\subsubsection{Comparison of Aleatoric Uncertainty Computed by Our Method, DDU and Logit-Sampling}
\begin{figure}
    \centering
    \begin{subfigure}{0.92\textwidth}
        \centering
        \includegraphics[width=\textwidth]{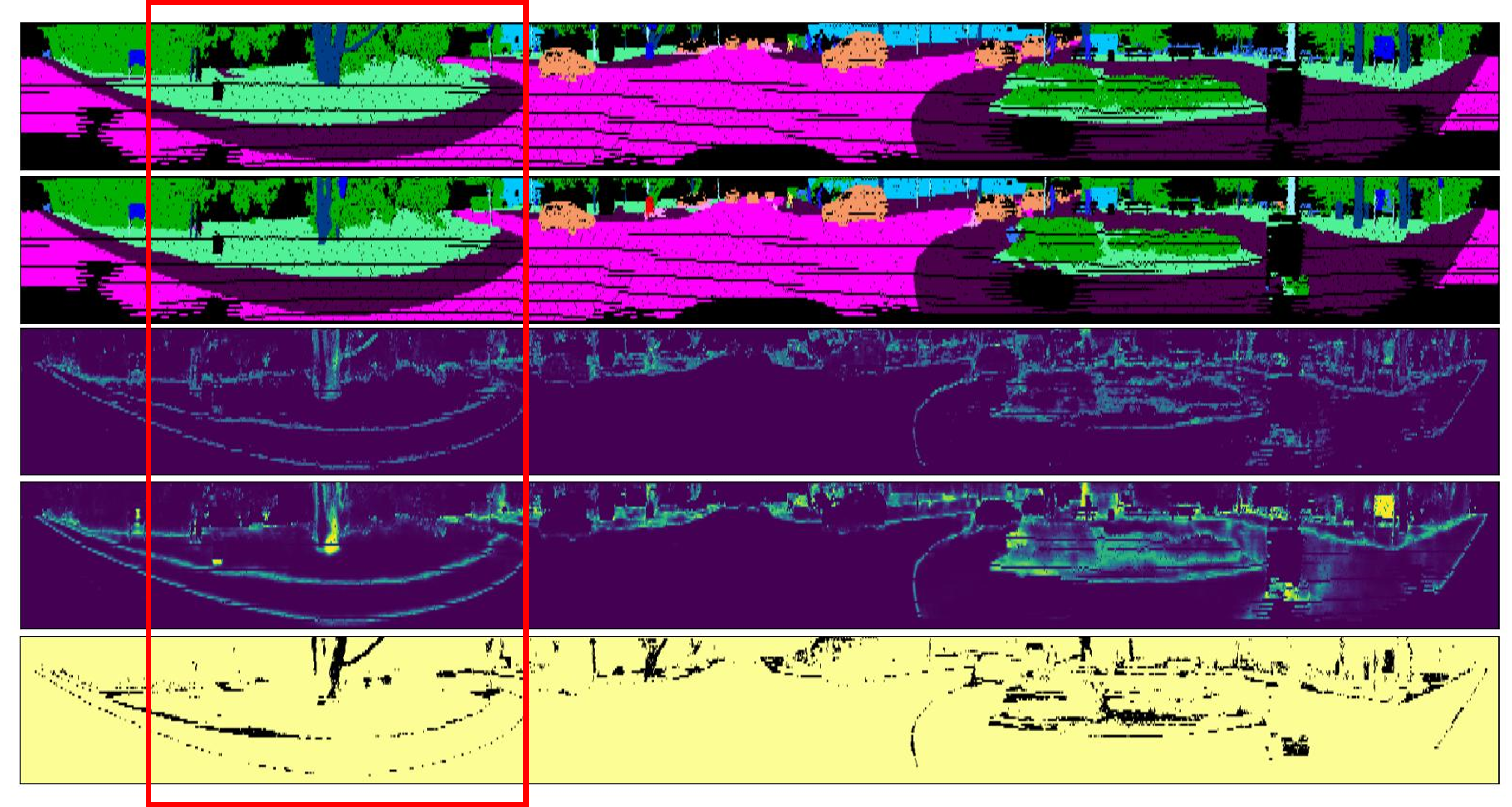}
        \caption{Our proposed method}
        \label{fig:AU_Comparison_a}
    \end{subfigure}
    \hfill
    \begin{subfigure}{0.9\textwidth}
        \centering
        \includegraphics[width=\textwidth]{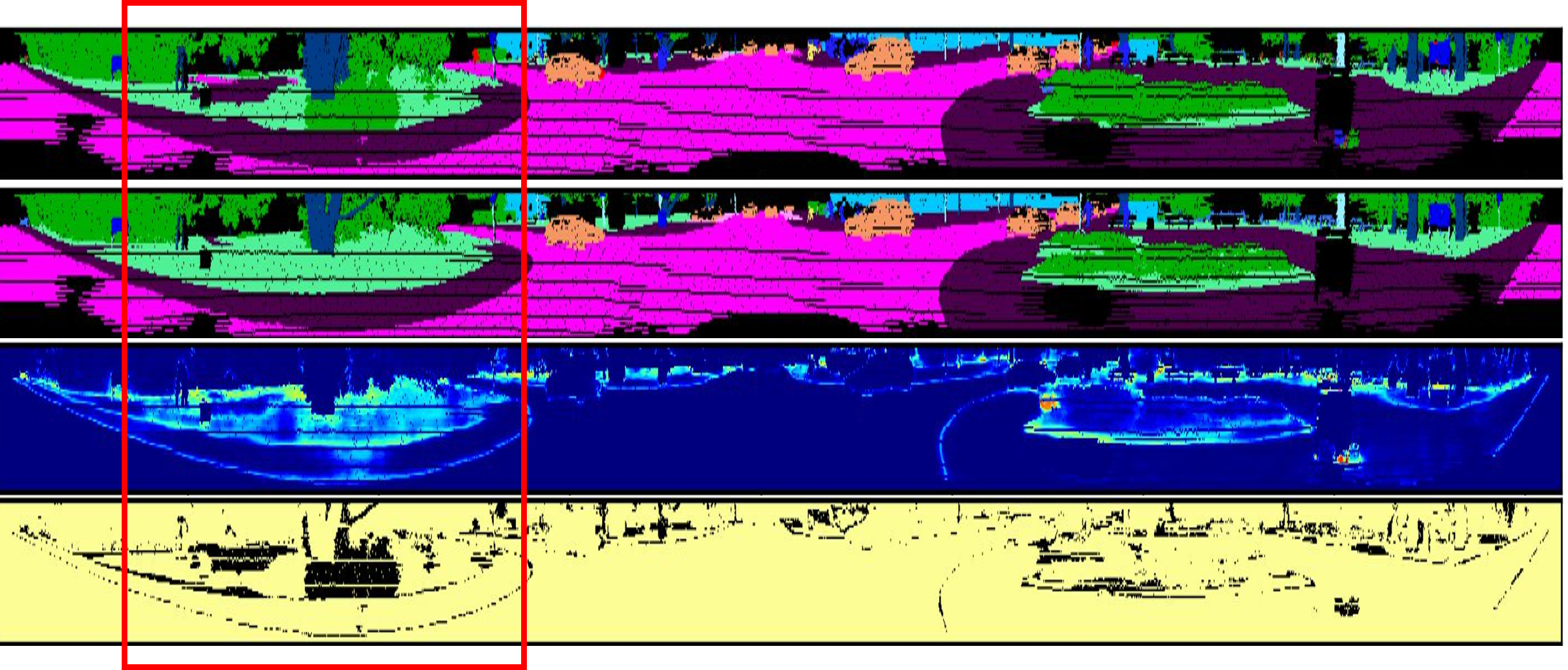}
        \caption{Logit-sampling method}
        \label{fig:AU_Comparison_b}
    \end{subfigure}
    \caption{Comparison of aleatoric uncertainty computed by \emph{(a) our proposed method} and \emph{(b) logit-sampling}. The figure displays, from top to bottom: \emph{ground truth map}, \emph{predicted map}, \emph{aleatoric uncertainty}, and \emph{error map}. In (a), the first aleatoric map shows our proposed method, and the second shows the DDU approach. The \emph{red box} highlights the limitations of logit-sampling, which misclassifies the terrain, leading to high errors and elevated aleatoric uncertainty. In contrast, our method demonstrates improved classification and more accurate class distinction. Comparing the aleatoric maps in (a) shows that our method aligns closely with the error map, while the DDU approach fails to detect high uncertainty in the trunk area highlighted in the red box.}
    \label{fig:AU_Comparison}

\end{figure}

\cref{fig:AU_Comparison} illustrates the comparison between aleatoric uncertainty computed as the entropy of responsibility values from our proposed method and that derived from the logit-sampling method. Additionally, \cref{fig:AU_Comparison_a} presents a comparison between our proposed aleatoric uncertainty map and the aleatoric uncertainty computed using the entropy of softmax outputs, as proposed by the DDU approach \cite{mukhoti2023}. 
The logit-sampling method, as shown in \cref{fig:AU_Comparison_b}, highlights many ambiguous areas, leading to classification errors visible in the error map (last row). The entire terrain, marked by a red box, exhibits high aleatoric uncertainty with logit-sampling (third map, \cref{fig:AU_Comparison_b}). In contrast, our proposed method, depicted in \cref{fig:AU_Comparison_a}, correctly identifies this terrain, distinguishing it from the surrounding vegetation and sidewalk. This discrepancy likely arises because the logit-sampling method introduces many additional trainable parameters, doubling those in the final layer to compute the output distribution directly.

Our method, shown in \cref{fig:AU_Comparison_a}, results in fewer errors, as evidenced by the error map. When comparing the aleatoric uncertainty maps in \cref{fig:AU_Comparison_a} (third map for our method, and fourth map for DDU), both methods primarily indicate high uncertainty at class boundaries. However, our method is more sensitive to ambiguous parts, highlighted for a trunk, which can be confused with vegetation (red box). We believe this uncertainty arises from mislabeled points that are difficult to distinguish between the trunk of a tree and its vegetation. In contrast, the DDU aleatoric uncertainty map fails to detect this uncertain and misclassified area.

\begin{figure}
    \centering
    \begin{subfigure}{0.30\textwidth}
        \centering
        \includegraphics[width=\textwidth]{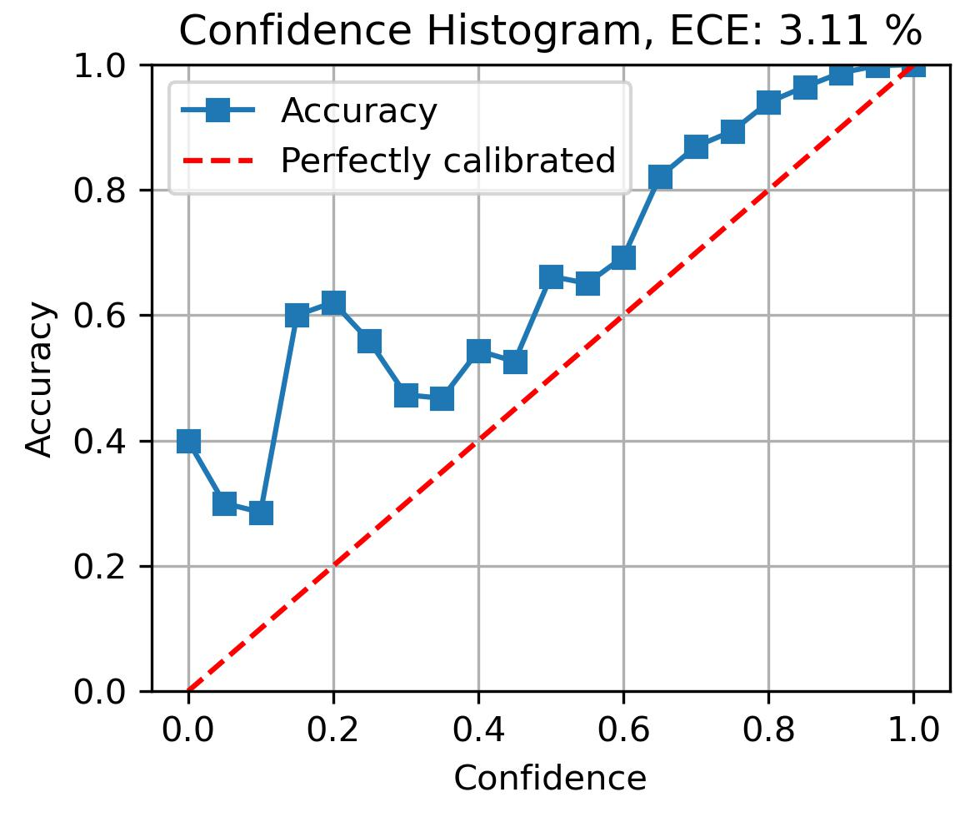}
        \caption{Our proposed method}
        \label{fig:conf_entresp}
    \end{subfigure}
    \begin{subfigure}{0.295\textwidth}
        \centering
        \includegraphics[width=\textwidth]{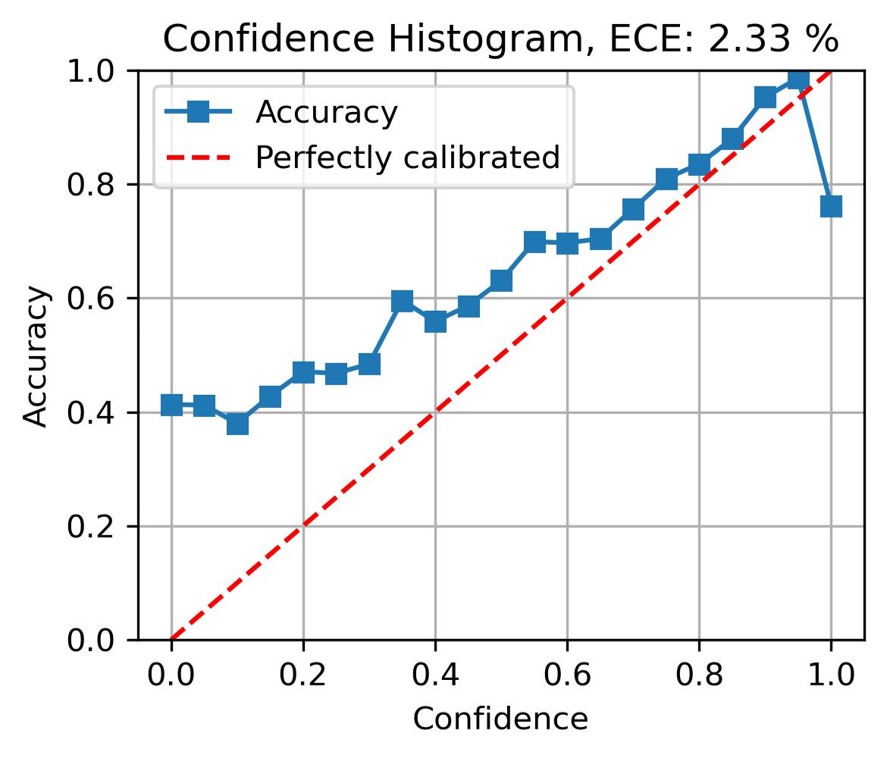}
        \caption{DDU method}
        \label{fig:000016_SN}
    \end{subfigure}
    \begin{subfigure}{0.30\textwidth}
        \centering
        \includegraphics[width=\textwidth]{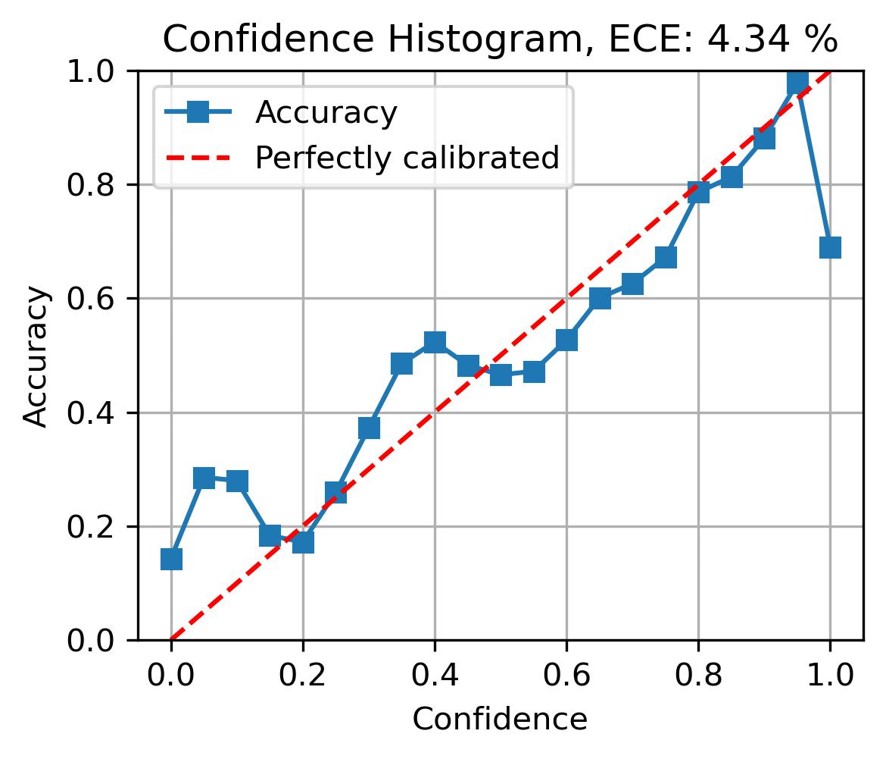}
        \caption{Logit-sampling method}
        \label{fig:000016_LS}
    \end{subfigure}
    \caption{Comparison of aleatoric uncertainty estimation using our method, DDU, and logit-sampling. Our method achieves an ECE of 3.11\% from raw responsibility values, while DDU achieves 2.33\% after temperature scaling, both outperforming logit-sampling with ECE of 4.34\%.}

    \label{fig:AU_Comparison_Conf}
\end{figure}

To quantitatively evaluate aleatoric uncertainty from our proposed metric, DDU, and logit-sampling approaches, we present confidence histograms and ECE metrics in \cref{fig:AU_Comparison_Conf}. The results show ECE values of 2.33\%, 3.11\%, and 4.34\% for softmax outputs (with temperature scaling), responsibility values, and logit-sampling, respectively. Although softmax outputs demonstrate better calibration due to temperature scaling, our approach shows superior qualitative performance, as the responsibility values are raw values while softmax outputs are calibrated by temperature scaling.

\section{Conclusion}

In this study, we introduce a method to distinguish OOD from ID samples and quantify both epistemic and aleatoric uncertainties in LiDAR scene semantic segmentation using a GMM in the feature space of a trained deep network. Our method eliminates the need for retraining multiple networks, relying instead on the uncertainty of a generative classifier (GMM) to compute epistemic uncertainty based on the variability in classification outputs across multiple GMMs. For aleatoric uncertainty, we propose using responsibility values from the GMMs, which better capture uncertainty by considering the entire data distribution rather than relying on softmax outputs.  

While OOD samples exhibit the highest epistemic uncertainty in our proposed epistemic maps, we introduced a separate OOD detection approach that directly identifies OOD samples and aligns with these high-uncertainty regions. This approach, based on the distance of samples to the Gaussian components of the GMM, operates independently of epistemic uncertainty values and does not require dedicated OOD samples during training, unlike other methods.

Our method is adaptable to any pre-trained model by considering a generative classifier (GMM) on its feature space, making it feasible for high-dimensional settings. However, assuming that class features follow a Multivariate Gaussian distribution may oversimplify deep network feature spaces. We are also uncertain if a single Gaussian component suffices or if multiple components might better approximate each class. Additionally, aleatoric uncertainty is computed from uncalibrated responsibility values, which may limit effectiveness. These issues will be addressed in future work.

We envision that our method will enhance LiDAR 3D point classification, improve OOD detection, and refine uncertainty quantification, increasing the reliability of 3D scene understanding for autonomous driving.

\section{Acknowledgement}
This project is supported by the German Research Foundation (DFG), as part of the Research Training Group i.c.sens,
GRK 2159, ‘Integrity and Collaboration in Dynamic Sensor
Networks’.

%
%
\bibliographystyle{splncs04}
\bibliography{main}

\begin{thebibliography}{10}
\providecommand{\url}[1]{\texttt{#1}}
\providecommand{\urlprefix}{URL }
\providecommand{\doi}[1]{https://doi.org/#1}

\bibitem{behley2019}
Behley, J., Garbade, M., Milioto, A., Quenzel, J., Behnke, S., Stachniss, C.,
  Gall, J.: Semantickitti: A dataset for semantic scene understanding of lidar
  sequences. In: Proceedings of the IEEE/CVF international conference on
  computer vision. pp. 9297--9307 (2019)

\bibitem{berman2018}
Berman, M., Triki, A.R., Blaschko, M.B.: The lov{\'a}sz-softmax loss: A
  tractable surrogate for the optimization of the intersection-over-union
  measure in neural networks. In: Proceedings of the IEEE conference on
  computer vision and pattern recognition. pp. 4413--4421 (2018)

\bibitem{bishop2006}
Bishop, C.M., Nasrabadi, N.M.: Pattern recognition and machine learning,
  vol.~4. Springer (2006)

\bibitem{cortinhal2020}
Cortinhal, T., Tzelepis, G., Erdal~Aksoy, E.: Salsanext: Fast,
  uncertainty-aware semantic segmentation of lidar point clouds. In: Advances
  in Visual Computing: 15th International Symposium, ISVC 2020, San Diego, CA,
  USA, October 5--7, 2020, Proceedings, Part II 15. pp. 207--222. Springer
  (2020)

\bibitem{dreissig2023}
Dreissig, M., Piewak, F., Boedecker, J.: On the calibration of uncertainty
  estimation in lidar-based semantic segmentation. In: 2023 IEEE 26th
  International Conference on Intelligent Transportation Systems (ITSC). pp.
  4798--4805. IEEE (2023)

\bibitem{gal2016}
Gal, Y., Ghahramani, Z.: Dropout as a bayesian approximation: Representing
  model uncertainty in deep learning. In: international conference on machine
  learning. pp. 1050--1059. PMLR (2016)

\bibitem{gawlikowski2023}
Gawlikowski, J., Tassi, C.R.N., Ali, M., Lee, J., Humt, M., Feng, J., Kruspe,
  A., Triebel, R., Jung, P., Roscher, R., et~al.: A survey of uncertainty in
  deep neural networks. Artificial Intelligence Review  \textbf{56}(Suppl 1),
  1513--1589 (2023)

\bibitem{gnanadesikan1972}
Gnanadesikan, R., Kettenring, J.R.: Robust estimates, residuals, and outlier
  detection with multiresponse data. Biometrics pp. 81--124 (1972)

\bibitem{guo2017}
Guo, C., Pleiss, G., Sun, Y., Weinberger, K.Q.: On calibration of modern neural
  networks. In: International conference on machine learning. pp. 1321--1330.
  PMLR (2017)

\bibitem{hullermeier2021}
H{\"u}llermeier, E., Waegeman, W.: Aleatoric and epistemic uncertainty in
  machine learning: An introduction to concepts and methods. Machine learning
  \textbf{110}(3),  457--506 (2021)

\bibitem{kendall2017}
Kendall, A., Gal, Y.: What uncertainties do we need in bayesian deep learning
  for computer vision? Advances in neural information processing systems
  \textbf{30} (2017)

\bibitem{lakshminarayanan2017}
Lakshminarayanan, B., Pritzel, A., Blundell, C.: Simple and scalable predictive
  uncertainty estimation using deep ensembles. Advances in neural information
  processing systems  \textbf{30} (2017)

\bibitem{lee2018}
Lee, K., Lee, K., Lee, H., Shin, J.: A simple unified framework for detecting
  out-of-distribution samples and adversarial attacks. Advances in neural
  information processing systems  \textbf{31} (2018)

\bibitem{liu2020}
Liu, J., Lin, Z., Padhy, S., Tran, D., Bedrax~Weiss, T., Lakshminarayanan, B.:
  Simple and principled uncertainty estimation with deterministic deep learning
  via distance awareness. Advances in neural information processing systems
  \textbf{33},  7498--7512 (2020)

\bibitem{loquercio2020}
Loquercio, A., Segu, M., Scaramuzza, D.: A general framework for uncertainty
  estimation in deep learning. IEEE Robotics and Automation Letters
  \textbf{5}(2),  3153--3160 (2020)

\bibitem{malinin2018}
Malinin, A., Gales, M.: Predictive uncertainty estimation via prior networks.
  Advances in neural information processing systems  \textbf{31} (2018)

\bibitem{maronna2002}
Maronna, R.A., Zamar, R.H.: Robust estimates of location and dispersion for
  high-dimensional datasets. Technometrics  \textbf{44}(4),  307--317 (2002)

\bibitem{milioto2019}
Milioto, A., Vizzo, I., Behley, J., Stachniss, C.: Rangenet++: Fast and
  accurate lidar semantic segmentation. In: 2019 IEEE/RSJ international
  conference on intelligent robots and systems (IROS). pp. 4213--4220. IEEE
  (2019)

\bibitem{mukhoti2023}
Mukhoti, J., Kirsch, A., van Amersfoort, J., Torr, P.H., Gal, Y.: Deep
  deterministic uncertainty: A new simple baseline. In: Proceedings of the
  IEEE/CVF Conference on Computer Vision and Pattern Recognition. pp.
  24384--24394 (2023)

\bibitem{nguyen2022}
Nguyen, V.L., Shaker, M.H., H{\"u}llermeier, E.: How to measure uncertainty in
  uncertainty sampling for active learning. Machine Learning  \textbf{111}(1),
  89--122 (2022)

\bibitem{ovadia2019}
Ovadia, Y., Fertig, E., Ren, J., Nado, Z., Sculley, D., Nowozin, S., Dillon,
  J., Lakshminarayanan, B., Snoek, J.: Can you trust your model's uncertainty?
  evaluating predictive uncertainty under dataset shift. Advances in neural
  information processing systems  \textbf{32} (2019)

\bibitem{van2021}
Van~Amersfoort, J., Smith, L., Jesson, A., Key, O., Gal, Y.: On feature
  collapse and deep kernel learning for single forward pass uncertainty. arXiv
  preprint arXiv:2102.11409  (2021)

\bibitem{van2020}
Van~Amersfoort, J., Smith, L., Teh, Y.W., Gal, Y.: Uncertainty estimation using
  a single deep deterministic neural network. In: International conference on
  machine learning. pp. 9690--9700. PMLR (2020)

\end{thebibliography}
\end{document}